%% file: main.tex
\documentclass[10pt,twocolumn,letterpaper]{article}

\usepackage{cvpr}              
\input{preamble}
\definecolor{cvprblue}{rgb}{0.21,0.49,0.74}
\usepackage[pagebackref,breaklinks,colorlinks,allcolors=cvprblue]{hyperref}

\usepackage{graphicx}
\usepackage{colortbl}
\usepackage{listings}
\usepackage{booktabs}
\usepackage{multirow}
\usepackage{array}
\usepackage{colortbl}
\usepackage{makecell}
\usepackage{quoting}
\usepackage{framed}
\usepackage{arydshln}
\usepackage[table]{xcolor}
\usepackage{longtable}
\usepackage{ltablex}
\usepackage[accsupp]{axessibility} 

\definecolor{tabblue}{HTML}{1f77b4}
\definecolor{taborange}{HTML}{ff7f0e}
\definecolor{tabgreen}{HTML}{2ca02c}
\definecolor{tabred}{HTML}{d62728}
\definecolor{tabpurple}{HTML}{9467bd}
\definecolor{tabpink}{HTML}{ff0080}

\def\paperID{} 
\def\confName{CVPR}
\def\confYear{2026}

\title{Exploring the Underwater World Segmentation without Extra Training}

\author{
Bingyu Li\textsuperscript{1,2}\thanks{Work done during an internship at TeleAI.} \quad
Tao Huo\textsuperscript{2,3} \quad
Da Zhang\textsuperscript{2,3} \quad
Zhiyuan Zhao\textsuperscript{2} \quad
Junyu Gao\textsuperscript{2} \quad
Xuelong Li\textsuperscript{2}\thanks{Corresponding Author} \\
\textsuperscript{1}University of Science and Technology of China, China \\
\textsuperscript{2}Institute of Artificial Intelligence (TeleAI), China Telecom, China \\
\textsuperscript{3}Northwestern Polytechnical University, China \\
}

\begin{document}
\maketitle
\input{sec/0_abstract}    
\input{sec/1_intro}
\input{sec/2_relawork}
\input{sec/3_dataset}
\input{sec/3_method}
\input{sec/4_experiment}
\input{sec/5_conclusion}
\paragraph{Acknowledgments}
This work was supported in part by the National Natural Science Foundation of China under Grants 62306241 and U62576284.

{
    \small
    \bibliographystyle{ieeenat_fullname}
    \bibliography{main}
}



\end{document}

%% file: sec/0_abstract.tex
\begin{abstract}
Accurate segmentation of marine organisms is vital for biodiversity monitoring and ecological assessment, yet existing datasets and models remain largely limited to terrestrial scenes. To bridge this gap, we introduce \textbf{AquaOV255}, the first large-scale and fine-grained underwater segmentation dataset containing 255 categories and over 20K images, covering diverse categories for open-vocabulary(OV) evaluation. Furthermore, we establish the first underwater OV segmentation benchmark, \textbf{UOVSBench}, by integrating AquaOV255 with five additional underwater datasets to enable comprehensive evaluation. Alongside, we present \textbf{Earth2Ocean}, a training-free OV segmentation framework that transfers terrestrial vision–language models (VLMs) to underwater domains without any additional underwater training. Earth2Ocean consists of two core components: a Geometric-guided Visual Mask Generator (\textbf{GMG}) that refines visual features via self-similarity geometric priors for local structure perception, and a Category-visual Semantic Alignment (\textbf{CSA}) module that enhances text embeddings through multimodal large language model reasoning and scene-aware template construction. Extensive experiments on the UOVSBench benchmark demonstrate that Earth2Ocean achieves significant performance improvement on average while maintaining efficient inference. The code and dataset are \href{https://github.com/LiBingyu01/Earth2Ocean}{\textcolor{tabpink}{Here}}\footnote{\textcolor{tabpink}{https://github.com/LiBingyu01/Earth2Ocean}}.
\end{abstract}

%% file: sec/1_intro.tex
\section{Introduction}
\label{sec:intro}

Detecting and segmenting marine organisms such as fish, corals, and other aquatic species are essential for marine monitoring, biodiversity assessment, and ecosystem management\cite{zheng2024marineinst,islam2020semantic, zheng2023marinegpt, haixin2023marinedet}. However, most existing segmentation datasets and models are developed for terrestrial or general-purpose visual scenes, leaving underwater environments largely underexplored\cite{hu2022overview}. The unique properties of underwater imagery—such as color attenuation, light scattering, low visibility, and blurred textures—further hinder the extraction of reliable visual features, rendering conventional segmentation methods inadequate for large-scale biological monitoring\cite{li2025maris}. 

\begin{figure}
\centering
\includegraphics[width=0.9\linewidth]{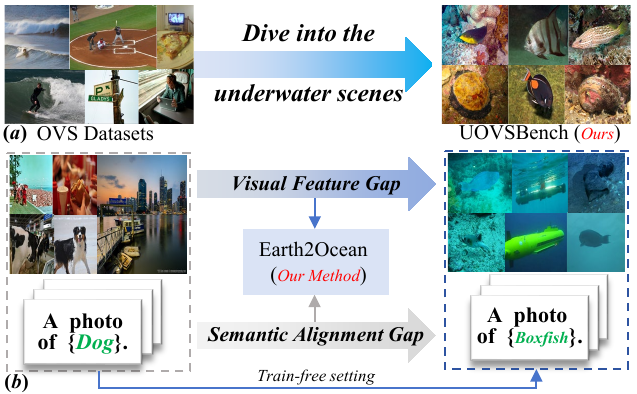} 
\caption{Our contributions: (a) proposing a fine-grained and diverse underwater dataset and benchmark, and (b) designing a practical training-free framework that enables zero-training adaptation for underwater scene transfer.}
\label{fig:pic_contribution}
\vspace{-10pt}
\end{figure}

Although several studies have extended visual understanding into underwater domains, they suffer from a crucial limitation: most existing works remain confined to \textit{closed-set recognition} \cite{li2025exploring,fu2023masnet}. For example, the USIS10K\cite{lian2024diving} dataset provides 10K underwater images but covers only ten predefined categories. Similarly, existing benchmarks \cite{ma2021underwater} often cover only a small number of predefined underwater categories, limiting fine-grained recognition across diverse marine organisms. Such coarse labeling severely limits the ecological applicability of these datasets. Moreover, most underwater segmentation models rely on extensive pretraining over a small set of categories\cite{fu2023masnet,qin2020u2}, resulting in computationally heavy pipelines and models restricted to predefined classes—further constraining their real-world utility. 

In this work, we address two above pressing challenges by (summarize in \cref{fig:pic_contribution}):  
\begin{enumerate}[leftmargin=*]
\item expanding the scope of underwater perception by constructing a fine-grained, multi-category dataset for comprehensive biological monitoring; and  
\item enhancing the practicality of underwater segmentation through a \textit{training-free} framework that enables direct transfer of terrestrial-based pretrained models to underwater environments without any extra underwater training.
\end{enumerate}

\begin{figure}
\centering
\includegraphics[width=0.9\linewidth]{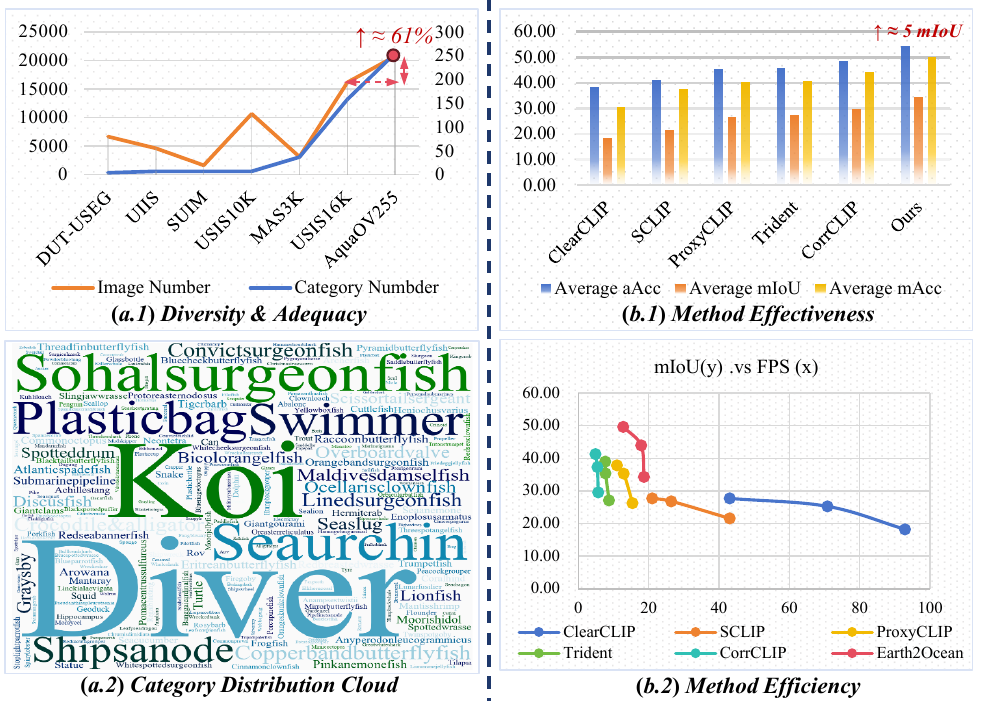}
\caption{(a.1) Compares image and category counts across datasets (AquaOV255 has about 61\% category growth); (a.2) Visualizes category distribution via word cloud; (b.1) Evaluates method effectiveness; (b.2) Trades off mIoU and FPS.}
\label{fig:pic_intro}
\vspace{-1.5em}
\end{figure}

To tackle the first challenge, we conducted an extensive review of existing underwater datasets and found that most contain fewer than 40 categories—insufficient for large-scale evaluation. To bridge this gap, we curated web-scale underwater imagery and developed a semi-automatic annotation pipeline based on Segment Anything (SAM) \cite{ISAT_with_segment_anything, kirillov2023segment, ravi2024sam2}, producing a fine-grained dataset with 255 categories (including background). To the best of our knowledge, this is \textit{the most detailed} underwater segmentation dataset to date (as shown in \cref{fig:pic_intro}(a.1-a.2)), encompassing a broad range of marine organisms and man-made objects. We also reorganized the existing datasets into an open-vocabulary format and, together with \textbf{AquaOV255}, established a comprehensive benchmark named \textbf{UOVSBench} for unified evaluation across diverse scenes.

To address the second challenge, we propose \textbf{Earth2Ocean}, a \textit{training-free} open-vocabulary segmentation framework that transfers the capabilities of terrestrial vision-language models (VLMs) to underwater scenarios. Earth2Ocean comprises two core components:

(1) \textit{Geometric-guided Visual Mask Generator} (\textbf{GMG}), which produces fine-grained, geometry-aware visual masks. In GMG, we leverage geometric-DINO features~\cite{zhang2022dino, yang2024depth, oquab2023dinov2} to compute self-similarity maps that serve as geometry-guided attention priors. Geometric information is relatively stable in underwater scenes~\cite{li2025maris} (verified by \cref{fig:pic_geometric_eff} and \cref{fig:pic_vis_segmap}(a)), and thus these priors are used to correct VLM visual features, resulting in visual masks that are better adapted to underwater structures.

(2) \textit{Category-visual Semantic Alignment} (\textbf{CSA}), which addresses \textit{the lack of underwater category–visual alignment} in conventional VLMs (e.g., CLIP~\cite{radford2021learning} and BLIP~\cite{li2022blip}) by introducing multimodal large language model (MLLM) reasoning\cite{bai2023qwen,yang2025qwen3,achiam2023gpt,team2023gemini,hurst2024gpt,li2023vision}. Specifically, we fuse the MLLM reasoning outputs, containing underwater scene awareness and category-visual alignment information, with the OV category list to enhance underwater semantic understanding, resulting in text features enriched with underwater knowledge and category-visual information (verified by \cref{fig:pic_vis_segmap}(b)).

The final segmentation results are obtained by integrating the outputs of GMG and CSA. Through this pipeline, Earth2Ocean achieves underwater-adapted mask generation and category-pixel prediction without any underwater training. As \cref{fig:pic_intro} (b.1-b.2), experiments on six datasets demonstrate that Earth2Ocean provides an average improvement of over 6 mIoU, while maintaining efficient inference suitable for real-world deployment.

In summary, our main contributions are as follows:
\begin{itemize}[leftmargin=*]
    \item \textbf{Dataset \& Benchmark:} AquaOV255 and UOVSBench provide comprehensive resources for open-vocabulary underwater segmentation.
    \item \textbf{Practical Framework:} Earth2Ocean transfers terrestrial VLMs to underwater scenarios for accurate segmentation without underwater training.
    \item \textbf{Complementary Modules:} GMG generates geometry-aware masks, and CSA leverages MLLM reasoning for better category–visual alignment.
\end{itemize}

%% file: sec/2_relawork.tex
\section{Related Work}
\label{sec:rela_work}

\begin{figure*}[htbp]
\centering
\includegraphics[width=0.9\linewidth]{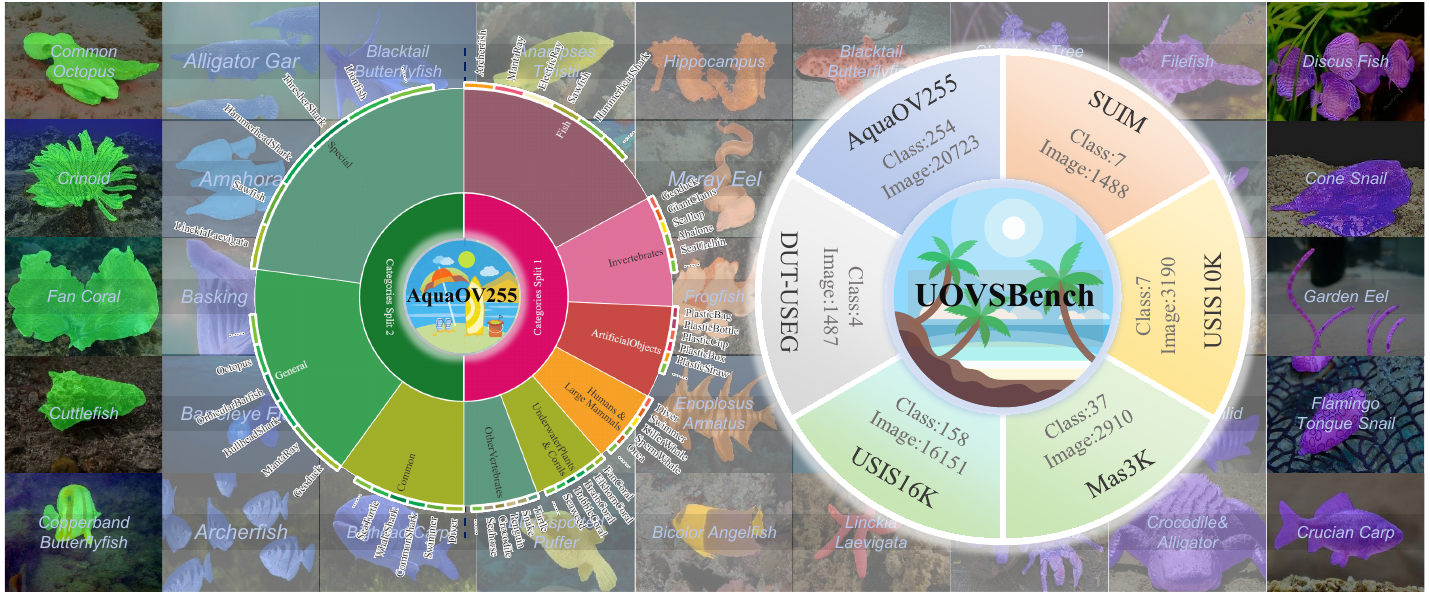}
\vspace{-1.1em}
\caption{\textbf{Foreground Part:} On the \texttt{left}, we show the AquaOV255 dataset along with its fine-grained splits; on the \texttt{right}, we present the UOVSBench Benchmark, including the datasets and image category counts. \textbf{Background Part:} Examples Visualization from Our \textbf{AquaOV255} Dataset. For further numerical analysis and properties, please refer to the Appendix.}
\vspace{-1.5em}
\label{fig:pic_dataset}
\end{figure*}

\paragraph{Underwater Segmentation.} has advanced with the introduction of dedicated benchmarks and deep learning methods. Early datasets such as SUIM~\cite{islam2020semantic}, UDD~\cite{liu2021new}, and DUT-USEG~\cite{ma2021underwater} established semantic understanding in aquatic scenes but were constrained in diversity and annotation quality. Subsequent datasets, including USOD10K~\cite{hong2023usod10k}, and MAS3K~\cite{li2020mas3k}, expanded scale and complexity, supporting more robust evaluation\cite{hong2025usis16k,lian2024diving, li2025uwsam}.  
Methodologically, FCN-, U-Net-, and GAN-based architectures were explored for underwater image restoration and segmentation, with WaterGAN~\cite{li2017watergan}, U-Net~\cite{ronneberger2015unet}, and adversarial approaches~\cite{patil2019motion} improving visual quality and segmentation robustness. More recent works incorporate attention and multi-scale fusion, e.g., U$^2$-Net~\cite{qin2020u2} and MASNet~\cite{fu2023masnet}, which better delineate object boundaries. Transformer-based models such as SegFormer~\cite{xie2021segformer} further improve generalization under variable visibility, while multimodal cues (e.g., depth, polarization) offer complementary information~\cite{hu2022overview}. 
Recently, the integration of VFMs into underwater tasks, particularly through SAM-driven approaches~\cite{li2025uwsam, lian2024diving, hong2024watersam}, reflects an emerging trend positioning VFMs as key enablers for underwater segmentation.

\paragraph{Training-free Open-Vocabulary Segmentation.} leverages pretrained vision-language models (VLMs) \cite{xie2025training,xie2026spatialqa, zhang2025cross} like CLIP~\cite{radford2021learning} for dense prediction without fine-tuning. Initial works such as MaskCLIP~\cite{zhou2022extract} mapped patch embeddings to textual features, while later refinements (CLIP Surgery~\cite{li2024explainability}, NACLIP~\cite{hajimiri2024pay}, SCLIP~\cite{wang2024sclip}) enhanced spatial fidelity via attention restructuring. Intermediate-layer exploitation (ITACLIP~\cite{aydin2024itaclip}, SC-CLIP~\cite{bai2025self}) and mask- or clustering-based regularization (CaR~\cite{sun2024cliprnn}, LaVG~\cite{kang2024lavg}) further improved robustness, revealing underused spatial cues in CLIP\cite{zhang2023tdec, zhang2023cnmbi, zheng2024deep}.  
Beyond pure CLIP-based solutions, researchers have incorporated auxiliary visual foundation models (VFMs) to inject stronger structural priors. ProxyCLIP~\cite{lan2024proxyclip} and CASS~\cite{kim2024cass} leverage DINO~\cite{caron2021emerging} for spatial priors, while DBA-CLIP~\cite{yang2024dbaclip}, Trident~\cite{shi2024harnessing} and CorrCLIP~\cite{zhang2025corrclip} pool embeddings guided by SAM~\cite{kirillov2023segment} or DINO masks. A parallel line of research investigates generative priors. Diffusion-based methods such as OVDiff~\cite{karazija2024ovdiff}, FreeDA~\cite{barsellotti2024freeda}, and FOSSIL~\cite{barsellotti2024fossil} construct visual or textual prototypes by synthesizing class-conditioned examples, while others extract cross-attention maps from generative models~\cite{wang2024diffseg, namekata2024emerdiff} to guide segmentation.

%% file: sec/3_dataset.tex
\section{Dataset: AquaOV255}
\label{sec:dataset}

\subsection{Data Collection}
The \textbf{AquaOV255} dataset was constructed by combining images crawled from public websites with existing underwater datasets such as USIS16K and USIS10K. We first defined 255 fine-grained categories across 6 super-classes and collected corresponding images using automated web crawling. All collected images were then cleaned to ensure high quality. In total, AquaOV255 contains 20,723 high-resolution underwater images spanning diverse scenes and object types.

\subsection{Annotation}
For annotation, we employed a semi-automatic workflow based on ISAT and SAM, involving five human experts to verify and correct labels. Specifically, large vision-language models assisted in category discrimination, while SAM enabled instance-level mask generation. Each image was carefully annotated to produce precise instance-level masks across all 255 categories.

\subsection{Dataset Analysis}
We analyze the basic characteristics of AquaOV255 in \cref{fig:pic_dataset} (left), it illustrates that AquaOV255 provides a rich and diverse collection of underwater scenes with fine-grained object categories, supporting research on open-vocabulary segmentation under challenging underwater conditions. It complements existing datasets by offering more detailed category coverage and high-quality instance annotations, making it suitable for both benchmarking and model development. Moreover, \cref{fig:pic_dataset} (background) demonstrates that our dataset contains a diverse range of underwater species and object categories.

\begin{figure*}[htbp]
\centering
\includegraphics[width=0.85\linewidth]{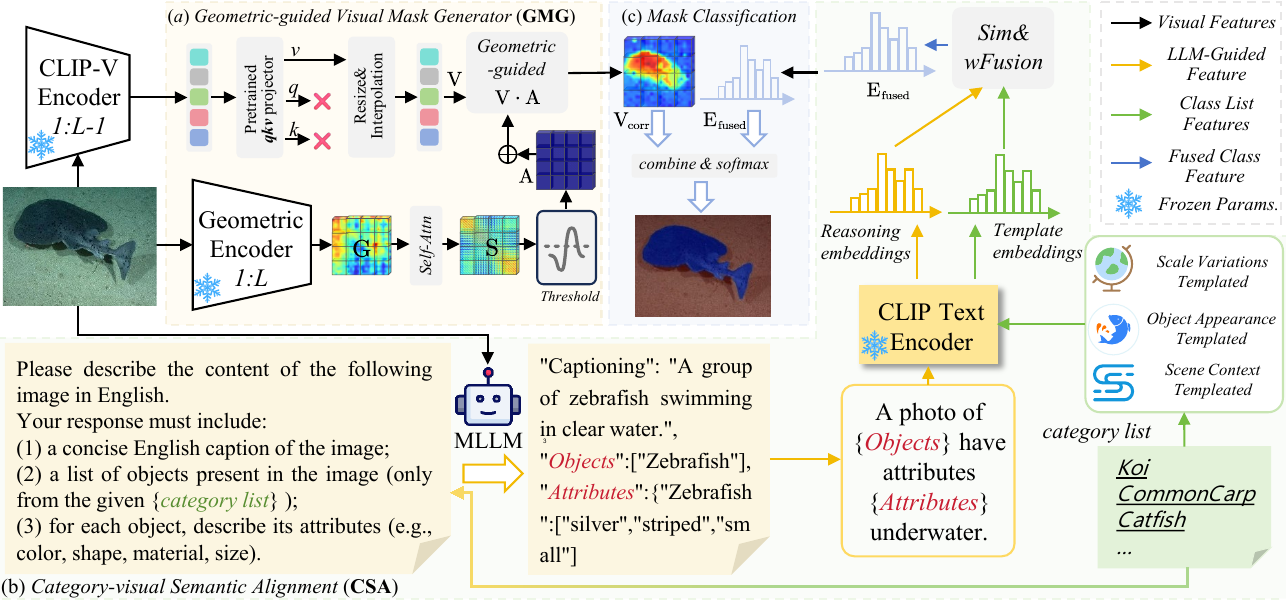}
\vspace{-1.1em}
\caption{This diagram outlines the \textbf{Earth2Ocean} framework, which integrates (a) \textit{Geometric-guided Visual Mask Generator} (\textbf{GMG}) to produce locally consistent and geometry-aware masks, (b) \textit{Category-visual Semantic Alignment} (\textbf{CSA}) leveraging MLLM reasoning to improve underwater category alignment, and (c) \textit{mask classification} for final pixel-level prediction.}
\vspace{-1.5em}
\label{fig:pic_method_main}
\end{figure*}

\section{Benchmark: UOVSBench}
\label{sec:UOVSBench}

\subsection{Benchmark Structure}
To systematically evaluate open-vocabulary segmentation (OVS) in underwater scenarios, as demonstrated in \cref{fig:pic_dataset} right part, we introduce \textbf{UOVSBench}. It consolidates five existing underwater segmentation datasets—USIS16K, SUIM, MAS3K, USIS10K, and DUT-USEG—by converting their category labels and image/mask pairs into an open-vocabulary segmentation format. Since these datasets often lack fine-grained category coverage, we further include our newly annotated \textbf{AquaOV255}, resulting in a comprehensive benchmark with diverse scenes. UOVSBench thus provides a standardized testbed for assessing model performance in challenging underwater conditions.

\subsection{Benchmark Evaluation Protocol}
We evaluate state-of-the-art training-free OVS models originally developed for terrestrial scenes under a \textit{training-free} setting, which emphasizes generalization and reflects practical deployment scenarios. All models are tested with OpenCLIP backbones of different scales (Base, Large, Huge). Other evaluation details, including preprocessing and metrics, are provided in the Appendix. The resulting performance metrics are summarized in ~\cref{tab:0_MainTable_1_UOVSBench}.

%% file: sec/3_method.tex
\section{Method: Earth2Ocean}
\label{method}

\subsection{Framework Overview}

Formally, given an input image $\mathbf{I} \in \mathbb{R}^{H \times W \times 3}$ and a set of textual concepts $\mathcal{C} = \{c_1, c_2, \dots, c_T\}$, the task is to predict a dense segmentation map $\mathbf{M}_{\text{pred}} \in \mathbb{R}^{T \times H \times W}$, such that each pixel is assigned to the most semantically compatible category in $\mathcal{C}$.

To tackle the challenging underwater open-vocabulary segmentation (UOVS) problem in a \textit{training-free} manner, our framework consists of three main components:

\begin{enumerate}
    \item \textbf{GMG}: extracts Geometric-aware visual features from VLM visual encoder.
    \item \textbf{CSA}: produces text embeddings enriched with underwater context and category-visual semantic alignment.
    \item \textbf{Mask Classification:} matches visual and text embeddings to perform pixel-wise classification.
\end{enumerate}

\subsection{Geometric-guided Visual Mask Generator}
\label{sec:GMG}

To address the unique challenges of underwater OV segmentation (UOVS) in a training-free manner, we refine the last-layer visual embeddings of CLIP to make them more spatially local, thereby enhancing its segmentation capability. We illustrate the mechanism in \cref{fig:pic_method_main}(a).

\subsubsection{Visual Encoding}
We first extract embeddings from the first $L-1$ layers of CLIP for an input image $\mathbf{I}$:
\begin{equation}
\mathbf{V} = \text{CLIP}_{1:L-1}(\mathbf{I}) \in \mathbb{R}^{N \times C}
\end{equation}
where $N = H \cdot W$ is the number of spatial positions and $C$ is the embedding dimension.  

\paragraph{Geometric Attention Maps.}  
The geometry encoder outputs multi-scale geometric features $\mathcal{G} = \{ \mathbf{G}_l \}_{l=0}^{{L_g}}, \mathbf{G}_l \in \mathbb{R}^{H_g \times W_g \times C_g},$
where $L_g$ denotes the total number of encoder stages. 
In our method, we adopt the geometric feature from the \textbf{last layer} ($\mathbf{G}_{L_g}$) of the encoder, which captures the most abstract and semantically rich geometric information (verified in \cref{fig:pic_ablation_layer}).

The last layer feature $\mathbf{G}_{L_g}$ is reshaped as $\hat{\mathbf{G}} \in \mathbb{R}^{C_g \times (H_g W_g)}$, and the geometric similarity map is computed as:
\begin{equation}
\mathbf{S} = \hat{\mathbf{G}}^\top \hat{\mathbf{G}} \in \mathbb{R}^{(H_g W_g) \times (H_g W_g)}.
\end{equation}

To enhance discriminative regions and suppress weak correlations, we apply mean-centering, scaling, and thresholding:
\begin{equation}
\tilde{\mathbf{S}} = \gamma (\mathbf{S} - \beta \overline{S}),
\end{equation}
\begin{equation}
\tilde{\mathbf{S}}_{ij} =
\begin{cases}
\tilde{\mathbf{S}}_{ij}, & \tilde{\mathbf{S}}_{ij} \ge 0,\\
-\infty, & \tilde{\mathbf{S}}_{ij} < 0,
\end{cases}
\end{equation}
\begin{equation}
\mathbf{A} = \mathrm{Softmax}(\tilde{\mathbf{S}}),
\end{equation}
where $\overline{S}$ denotes the mean of all elements in $\mathbf{S}$, and $\beta, \gamma$ are hyper-parameters controlling centering and scaling.

\paragraph{Geometric-guided Visual Correction.}  
The CLIP visual embeddings $\mathbf{V}$ are first interpolated to $\mathbf{V}' \in \mathbb{R}^{C \times (H_g W_g)}$ 
to match the spatial resolution of the geometric attention map.
The final attention map $\mathbf{A}$ is used to refine the spatially interpolated CLIP visual embeddings $\mathbf{V'}$:
\begin{equation}
\mathbf{V}_{\text{corr}} = \mathbf{A} \cdot \mathbf{V'}.
\end{equation}

\subsection{Category-Visual Semantic Alignment}

The primary goal of semantic enhancement is to enable the model to adapt to underwater \textit{scene context} and ensure \textit{category-visual alignment}. Scene context helps the model recognize that the target content appears in underwater environments, while category-visual alignment ensures that visual features are correctly associated with their corresponding textual descriptions. Our CSA module (\cref{fig:pic_method_main}(b)) is designed to address both aspects effectively.

\subsubsection{Underwater-Aware Scene Context Enhancement}

To provide the model with underwater scene information, we employ a template-based semantic enhancement. Underwater environments are characterized by low lighting and distinctive environmental semantics. To generate underwater-specific textual templates for training-free UOVS, we design a structured templates that enhances semantics from multiple aspects, including \texttt{object appearance, scene context, environmental conditions, interactions, and scale variations}. Collectively, these form $T$ templates for each category. The templates are instantiated dynamically by combining with the target category, producing underwater-aware textual embeddings $\mathbf{E}_t^1 \in \mathbb{R}^{T \times N \times C}$. We then average the embeddings across all templates to obtain a final representation $\mathbf{E}_t \in \mathbb{R}^{T \times C}$, where $T$ is the number of categories. These embeddings improve object recognition in a \textit{training-free} manner.

\begin{figure}[htbp]
\centering
\includegraphics[width=0.96\linewidth]{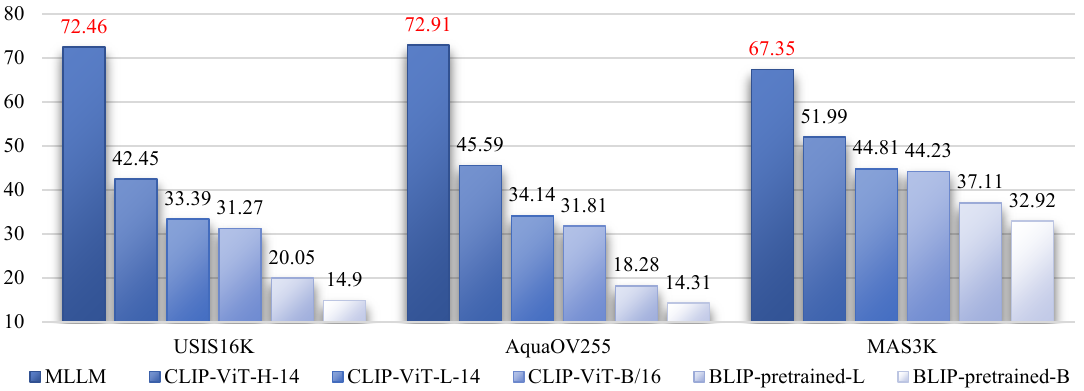}
\caption{MLLMs demonstrate superior capability over VLMs in capturing semantic and visual cues for underwater object classification accuracy (Y-axis).}
\label{fig:pic_classify_comparison}
\vspace{-10pt}
\end{figure}

\input{table/0_MainTable_1}

\subsubsection{MLLM-Guided Category-Visual Alignment}

While the Underwater-Aware Template Enhancement introduces rich scene context, it does not explicitly address the misalignment between category semantics and visual features. Such \textit{category-visual misalignment} can degrade recognition performance, particularly under challenging underwater conditions. As our preliminary classification experiments shown in \cref{fig:pic_classify_comparison}, although VLMs like CLIP~\cite{radford2021learning} and BLIP~\cite{li2022blip} achieve strong zero-shot recognition in earth scenarios, they struggle to discriminate between visually similar underwater categories. To bridge this gap and enable effective \textit{training-free} adaptation, we propose an \textbf{MLLM-Guided Category-Visual Alignment} mechanism, which injects high-level reasoning knowledge from a multimodal large language model (MLLM) into textual embeddings, enhancing their alignment with underwater visual features.

\paragraph{Reasoning-Aware Text Embedding.}  
Given an input image, the MLLM is prompted to generate:  
(1) a concise image caption,  
(2) a list of objects drawn from the category set, and  
(3) attribute descriptions for each object (e.g., color, shape, size).  
For instance, the MLLM may produce:

\vspace{-0.1in}
\begin{leftbar}
\noindent
{\small
\texttt{
\{\quad "Caption": "A group of zebrafish swimming in clear water.",\\
\quad "Objects": ["Zebrafish"],\\
\quad "Attributes": \{\\
\quad\quad "Zebrafish": ["silver", "striped", "small"]\\
\quad \}\}
}}
\end{leftbar}
\vspace{-0.1in}

We then combine each object with its attributes using a structured template to form reasoning sentences:  
\texttt{``A photo of \{Objects\} that have attributes \{Attributes\} underwater.''}  
Encoding these sentences with the CLIP text encoder produces reasoning-aware embeddings $\mathbf{E}_r \in \mathbb{R}^{1 \times C}$, which capture \textbf{instance-specific, attribute-rich, and domain-relevant semantics}. These embeddings significantly strengthen the alignment between textual and visual features in challenging underwater environments.

\paragraph{Similarity-Guided Embedding Fusion.}  
To enhance the original template embeddings $\mathbf{E}_t \in \mathbb{R}^{T \times C}$, we fuse them with a reasoning-aware embedding $\mathbf{E}_r \in \mathbb{R}^{1 \times C}$ using a similarity-guided approach. This is particularly important in underwater scenarios, where visual cues are often degraded or ambiguous. By selectively integrating reasoning-aware information, the model strengthens category-visual alignment and captures instance-specific, domain-relevant semantics.

First, both template and reasoning embeddings $\mathbf{E}_t, \mathbf{E}_r$ are $L_2$-normalized: $\mathbf{E}_t^\text{norm}, \mathbf{E}_r^\text{norm}$. The cosine similarities are computed and the corresponding fusion weights are defined element-wise:
\begin{equation}
\mathbf{s} = \mathbf{E}_t^\text{norm} \cdot \mathbf{E}_r^{\text{norm}^\top} \in \mathbb{R}^{T \times 1}
\end{equation}
\begin{equation}
\mathbf{w} = \min(\mathbf{s}, w_\text{max}) \odot (\mathbf{s} \ge \tau)
\end{equation}
where $\odot$ denotes element-wise multiplication with the indicator vector $(\mathbf{s} \ge \tau)$.  

Finally, the fused embeddings are computed in a single matrix operation:
\begin{equation}
\mathbf{E}_\text{fused} = \frac{\mathbf{E}_t + \mathbf{w} \cdot \mathbf{E}_r}{\|\mathbf{E}_t + \mathbf{w} \cdot \mathbf{E}_r\|_2}
\end{equation}

This produces attribute-rich and domain-relevant text embeddings, improving zero-shot matching with visual features without additional training.

\subsection{Visual Mask Classification}

As \cref{fig:pic_method_main}(c), given the visual features $\mathbf{V}_{\text{corr}} \in \mathbb{R}^{HW \times C}$ produced by \textbf{GMG} and the semantic embeddings $\mathbf{E}_\text{fused} \in \mathbb{R}^{T \times C}$ from \textbf{CSA}, the final prediction $\mathbf{M} \in \mathbb{R}^{T \times H \times W}$ is obtained via a linear projection along the channel dimension. Per-pixel classification probabilities are subsequently derived using a \texttt{softmax} operation.
\begin{equation}
\mathbf{M} = \mathbf{E}_\text{fused} \cdot \mathbf{V}_{\text{corr}}^\top
\end{equation}
\begin{equation}
\mathbf{M}_{\text{pred}} = \text{softmax}(\mathbf{M})
\end{equation} 

%% file: table/0_MainTable_1.tex
\begin{table*}[htbp]
\centering
\small
\renewcommand{\arraystretch}{1.2} 
\setlength{\tabcolsep}{2.5pt} 
\caption{Comparison of \textbf{Earth2Ocean} with previous training-free OVS models on \textbf{UOVSBench} in terms of aAcc, mIoU, and mAcc. Across three settings (ViT-B/16, ViT-L/14, and ViT-H/14), our method consistently outperforms the previous best-performing models. The best results are marked in \textcolor{tabred}{\textbf{red}} and the second-best in \textcolor{tabblue}{\textbf{blue}}.}.
\label{tab:0_MainTable_1_UOVSBench}
\resizebox{0.85\linewidth}{!}{%
\begin{tabular}{c|c|c|
ccc|ccc|ccc|ccc|ccc|ccc|ccc}
\toprule
\multirow{2}{*}{\textbf{Backbone}} & \multirow{2}{*}{\textbf{Method}} & \multirow{2}{*}{\textbf{Publication}}
& \multicolumn{3}{c}{\cellcolor{blue!5}DUT-Seg} 
& \multicolumn{3}{c}{\cellcolor{blue!5}MAS3K} 
& \multicolumn{3}{c}{\cellcolor{blue!5}SUIM} 
& \multicolumn{3}{c}{\cellcolor{blue!5}USIS10K} 
& \multicolumn{3}{c}{\cellcolor{blue!5}USIS16K} 
& \multicolumn{3}{c}{\cellcolor{blue!5}AquaOV255} 
& \multicolumn{3}{c}{\cellcolor{orange!5}Average} \\
\cmidrule(lr){4-6} \cmidrule(lr){7-9} \cmidrule(lr){10-12} \cmidrule(lr){13-15} 
\cmidrule(lr){16-18} \cmidrule(lr){19-21} \cmidrule(lr){22-24}
 & & & aAcc & mIoU & mAcc & aAcc & mIoU & mAcc & aAcc & mIoU & mAcc & aAcc & mIoU & mAcc 
 & aAcc & mIoU & mAcc & aAcc & mIoU & mAcc & aAcc & mIoU & mAcc \\
\midrule
\multirow{6}{*}{ViT-B/16} 
 & ClearCLIP & \textcolor{gray!50}{ECCV2024} & 18.10 & 5.55 & 25.25 & 43.54 & 23.11 & 31.84 & 62.57 & 26.08 & 42.98 & 54.01 & 26.59 & 37.04 & 29.31 & 16.78 & 26.29 & 20.93 & 11.29 & 18.16 & 38.08 & 18.23 & 30.26 \\
 & SCLIP     & \textcolor{gray!50}{ECCV2024} & 34.90 & 18.88 & 45.42 & 37.66 & 17.55 & 29.48 & 64.34 & 37.60 & 61.51 & 63.34 & 33.09 & 47.57 & 27.26 & 14.26 & 25.69 & 18.12 & 8.03 & 15.61 & 40.94 & 21.57 & 37.55 \\
 & ProxyCLIP & \textcolor{gray!50}{ECCV2024} & 24.74 & 13.24 & 35.74 & 44.76 & 24.05 & 36.65 & 73.07 & 48.47 & 63.40 & 68.86 & 39.10 & 51.08 & 35.34 & 20.45 & 32.66 & 25.00 & 12.44 & 21.60 & 45.30 & 26.29 & 40.19 \\
 & Trident   & \textcolor{gray!50}{ICCV2025} & 22.15 & 12.29 & 33.81 & 45.85 & 25.76 & 38.03 & 73.35 & 48.72 & 62.39 & 68.07 & 39.05 & 50.26 & 37.88 & 22.86 & 35.06 & 26.88 & 14.13 & 23.57 & 45.70 & 27.14 & 40.52 \\
 & CorrCLIP  & \textcolor{gray!50}{ICCV2025} & 27.24 & 16.81 & 36.41 & 48.26 & 27.19 & 41.55 & 73.87 & 51.00 & 68.05 & 67.56 & 40.39 & 53.15 & 42.44 & 25.95 & 39.54 & 30.57 & 16.02 & 26.54 & \textcolor{tabblue}{\textbf{48.32}} & \textcolor{tabblue}{\textbf{29.56}} & \textcolor{tabblue}{\textbf{44.21}} \\
\rowcolor{gray!10} & Earth2Ocean & \textcolor{gray!50}{--} & 52.69 & 34.07 & 53.64 & 50.42 & 28.41 & 42.34 & 73.06 & 51.97 & 72.73 & 71.85 & 44.63 & 59.24 & 45.42 & 29.02 & 43.03 & 32.61 & 17.81 & 28.06 & \textcolor{tabred}{\textbf{54.34}} & \textcolor{tabred}{\textbf{34.32}} & \textcolor{tabred}{\textbf{49.84}}\\

\midrule 
\multirow{6}{*}{ViT-L/14} 
 & ClearCLIP & \textcolor{gray!50}{ECCV2024} & 29.31 & 14.90 & 44.36 & 51.22 & 32.09 & 48.98 & 65.88 & 32.27 & 45.43 & 50.06 & 28.15 & 39.50 & 39.96 & 25.46 & 36.52 & 32.52 & 18.76 & 29.05 & 44.83 & 25.27 & 40.64 \\
 & SCLIP     & \textcolor{gray!50}{ECCV2024} & 55.66 & 33.38 & 58.80 & 46.50 & 25.33 & 41.66 & 56.72 & 34.20 & 60.27 & 62.19 & 33.04 & 46.98 & 33.84 & 20.46 & 32.71 & 26.21 & 14.56 & 23.75 & 46.85 & 26.83 & 44.03 \\
 & ProxyCLIP & \textcolor{gray!50}{ECCV2024} & 58.05 & 32.16 & 58.83 & 54.79 & 33.30 & 53.10 & 72.37 & 51.65 & 66.01 & 65.43 & 41.99 & 54.29 & 45.23 & 30.55 & 42.35 & 37.28 & 22.29 & 33.03 & 55.53 & 35.32 & 51.27 \\
 & Trident   & \textcolor{gray!50}{ICCV2025} & 54.10 & 29.41 & 56.54 & 54.96 & 34.25 & 54.53 & 72.28 & 51.47 & 64.66 & 65.56 & 41.68 & 53.42 & 47.22 & 32.13 & 43.86 & 38.70 & 23.44 & 34.29 & 55.47 & 35.40 & 51.22 \\
 & CorrCLIP  & \textcolor{gray!50}{ICCV2025} & 62.93 & 37.47 & 58.53 & 56.97 & 34.57 & 54.01 & 74.01 & 54.95 & 71.38 & 64.37 & 42.81 & 57.38 & 46.99 & 31.32 & 43.59 & 38.54 & 22.85 & 34.12 & \textcolor{tabblue}{\textbf{57.30}} & \textcolor{tabblue}{\textbf{37.33}} & \textcolor{tabblue}{\textbf{53.17}} \\
\rowcolor{gray!10}   & Earth2Ocean & \textcolor{gray!50}{--} & 68.28 & 41.32 & 61.86 & 63.99 & 40.94 & 61.87 & 74.27 & 55.17 & 75.81 & 70.36 & 46.90 & 61.45 & 59.97 & 45.13 & 58.04 & 52.59 & 34.53 & 47.74 & \textcolor{tabred}{\textbf{64.91}} & \textcolor{tabred}{\textbf{44.00}} & \textcolor{tabred}{\textbf{61.13}}\\

\midrule
\multirow{6}{*}{ViT-H/14} 
 & ClearCLIP & \textcolor{gray!50}{ECCV2024} & 22.69 & 13.23 & 41.83 & 60.53 & 39.26 & 57.64 & 65.21 & 28.41 & 43.88 & 47.83 & 22.80 & 31.80 & 51.34 & 36.08 & 49.07 & 40.75 & 26.42 & 37.93 & 48.06 & 27.70 & 43.69 \\
 & SCLIP     & \textcolor{gray!50}{ECCV2024} & 53.60 & 30.80 & 57.53 & 47.95 & 26.14 & 43.11 & 60.57 & 34.43 & 59.04 & 56.09 & 33.03 & 49.84 & 38.90 & 25.32 & 39.92 & 28.72 & 16.58 & 27.14 & 47.64 & 27.72 & 46.10 \\
 & ProxyCLIP & \textcolor{gray!50}{ECCV2024} & 50.86 & 31.29 & 55.21 & 64.81 & 37.91 & 56.33 & 75.11 & 52.34 & 67.62 & 70.93 & 41.26 & 50.43 & 53.12 & 37.89 & 52.42 & 41.32 & 26.41 & 38.88 & 59.36 & 37.85 & 53.48 \\
 & Trident   & \textcolor{gray!50}{ICCV2025} & 42.73 & 27.15 & 51.87 & 64.08 & 40.43 & 58.34 & 75.76 & 52.83 & 66.39 & 71.05 & 40.96 & 49.56 & 57.94 & 42.43 & 56.46 & 45.61 & 30.17 & 43.04 & 59.53 & 39.00 & 54.28 \\
 & CorrCLIP  & \textcolor{gray!50}{ICCV2025} & 54.15 & 36.10 & 55.44 & 62.43 & 40.34 & 57.73 & 76.20 & 54.34 & 70.84 & 75.04 & 44.88 & 54.27 & 57.82 & 42.35 & 56.01 & 44.86 & 29.57 & 41.56 & \textcolor{tabblue}{\textbf{61.75}} & \textcolor{tabblue}{\textbf{41.26}} & \textcolor{tabblue}{\textbf{55.98}}\\
\rowcolor{gray!10}   &Earth2Ocean & \textcolor{gray!50}{--} & 74.37 & 55.24 & 67.66 & 67.26 & 47.15 & 67.40 & 78.34 & 61.04 & 78.85 & 72.62 & 49.12 & 62.04 & 60.45 & 45.68 & 59.99 & 55.98 & 39.76 & 53.37  & \textcolor{tabred}{\textbf{68.17}} & \textcolor{tabred}{\textbf{49.67}} & \textcolor{tabred}{\textbf{64.89}} \\
\bottomrule
\end{tabular}%
}
\vspace{-1.2em}
\end{table*}

%% file: sec/4_experiment.tex
\section{Experiments And Results}
\label{sec:experiments_results}
\subsection{Experimental Details}
All experiments were conducted on a single NVIDIA RTX 4090 GPU. Our implementation is based on \texttt{MMSegmentation}, and the geometric encoder is adopted from prior works~\cite{zhang2022dino, yang2024depth, oquab2023dinov2}. $L_g$ correspond to the \{0,1,2,3\}-th stage, respectively. All compared method are reproduced on the same hardware for fair evaluation, the details are in the Appendix. To improve inference efficiency, we save the MLLM inference information in \texttt{JSON} format. During the \texttt{init} stage, this information is pre-encoded into the CLIP text encoder, enabling fast, end-to-end inference. 
Our default parameters are set as $\beta = 1.2$, $\gamma = 3.0$, $w_\text{max} = 0.5$, and $\tau = 0.5$, and the ablation experiments of the model are conducted using the \texttt{ViT-B/16} model.

\input{table/0_MainTable_2}
\subsection{Comparisons with State-of-the-Art Methods}
We conducted two sets of experiments: a comprehensive evaluation on UOVSBench and a fine-grained analysis on AquaOV255 to validate the effectiveness of our Earth2Ocean model.

\paragraph{Comprehensive evaluation on UOVSBench.}  
As shown in \cref{tab:0_MainTable_1_UOVSBench}, Earth2Ocean consistently outperforms previous training-free OVS models across all backbones (ViT-B/16, ViT-L/14, and ViT-H/14) and datasets. In terms of average metrics, our model improves aAcc, mIoU, and mAcc by notable margins compared to the previous best-performing methods, demonstrating robust generalization across diverse underwater scenarios. The gains are especially significant for the larger backbones, indicating that Earth2Ocean effectively leverages richer visual representations for open-vocabulary segmentation.

\paragraph{Fine-grained analysis on AquaOV255.}  
\cref{tab:0_MainTable_2_categories} presents category-wise mIoU and mAcc on the AquaOV255 dataset, which contains 255 fine-grained underwater categories grouped by taxonomy and commonness. Earth2Ocean consistently achieves higher performance across all categories, including artificial objects, invertebrates, fishes, and corals. The improvements are particularly pronounced in less common and special categories, highlighting the model's ability to capture rare and challenging classes. This detailed analysis demonstrates that Earth2Ocean not only excels in overall segmentation accuracy but also maintains strong performance at the category level, effectively addressing the long-tail distribution in underwater open-vocabulary segmentation.


\subsection{Ablation Study of the Methods}
\paragraph{Ablation Study of Components.}  
We evaluate the contribution of each component in our model as shown in \cref{tab:performance_comparison_conponent}. The methods are complementary, with each component contributing to improved feature alignment and cross-modal interaction.
\paragraph{Ablation Study of Hyperparameters.}  
\cref{tab:hyperparam_ablation_transpose} presents the effects of different hyperparameters on overall performance. We find that the model achieves the best results when $\beta=1.2$, $\gamma=3.0$, $w_\text{max}=0.5$, and $\tau=0.1$, indicating a stable and optimal configuration for feature weighting, alignment strength, and attention thresholds.
\input{table/1_ablation_component}
\input{table/2_ablation_hyper_parameter}

\begin{figure}[htbp]
\centering
\includegraphics[width=\linewidth]{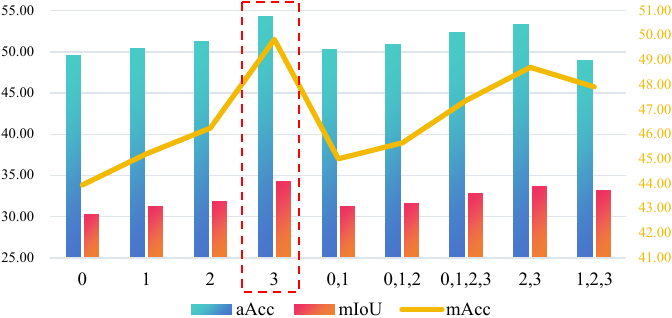}
\caption{Ablation study of different geometric feature layer.}
\label{fig:pic_ablation_layer}
\end{figure}

\paragraph{Ablation Study of Geometric Feature Layer.}  
As illustrated in \cref{fig:pic_ablation_layer}, we analyze the contribution of geometric features from different layers. The results indicate that using only the last layer (stage 3) yields the best performance, suggesting that high-level geometric information is most beneficial for feature alignment in our framework.

\subsection{Analysis of the Impact of Different MLLMs}  
As shown in \cref{tab:Earth2Ocean_ViTB_avg}, Earth2Ocean's segmentation accuracy relies heavily on the intrinsic multimodal capabilities of the chosen MLLM, with GPT-4o consistently outperforming Qwen variants regardless of model scale.

\input{table/4_analysis_mllms}

\subsection{Model Efficiency Comparison}  
\cref{tab:flops_comparison_fps} compares memory, FLOPs, parameters, FPS, and segmentation accuracy. Using pre-encoded MLLM features in the \texttt{init} stage, our Earth2Ocean model achieves the best balance between efficiency and accuracy.
\input{table/3_analysis_memory_time}

\begin{figure}[htbp]
\centering
\includegraphics[width=\linewidth]{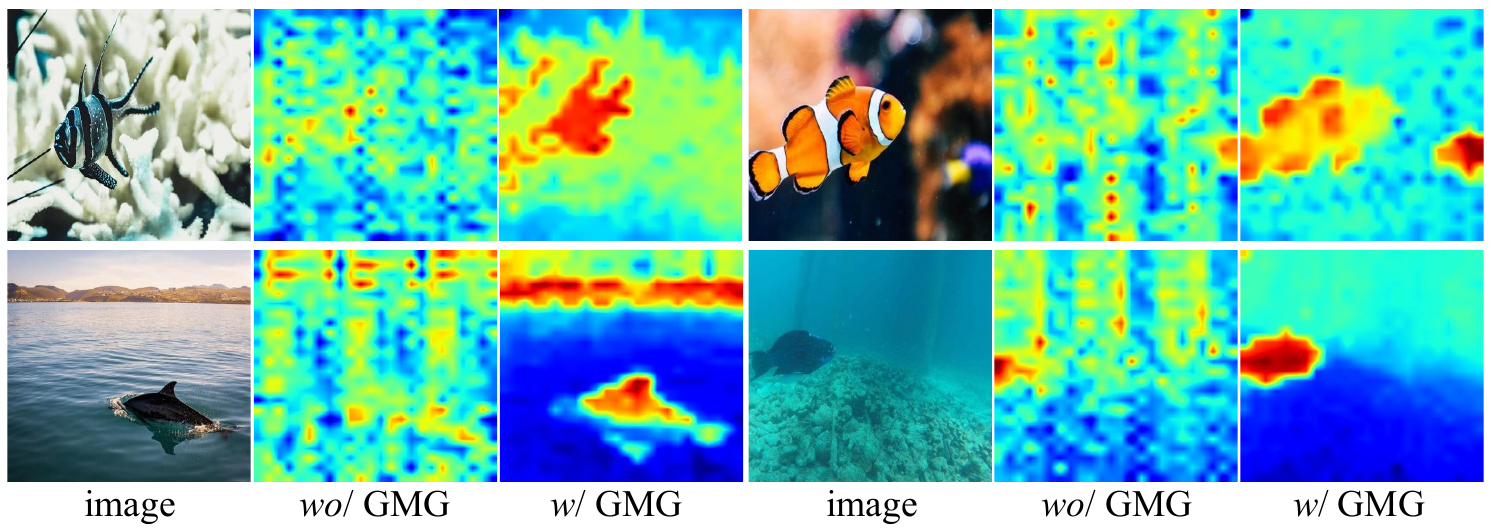}
\vspace{-15pt}
\caption{Visualization study of GMG effectiveness, showing visual feature maps with and without the GMG module.}
\label{fig:pic_geometric_eff}
\vspace{-12pt}
\end{figure}

\subsection{Model Visualization Comparison}

\paragraph{Visualization of GMG Effectiveness.}
As shown in \cref{fig:pic_geometric_eff}, the GMG-enhanced heatmaps accurately highlight underwater target objects, such as fish and corals, whereas the heatmaps without GMG exhibit more scattered and noisy activations, demonstrating the module’s effectiveness in calibrating geometric features in challenging underwater scenes.

\begin{figure}[ht]
\centering
\includegraphics[width=\linewidth]{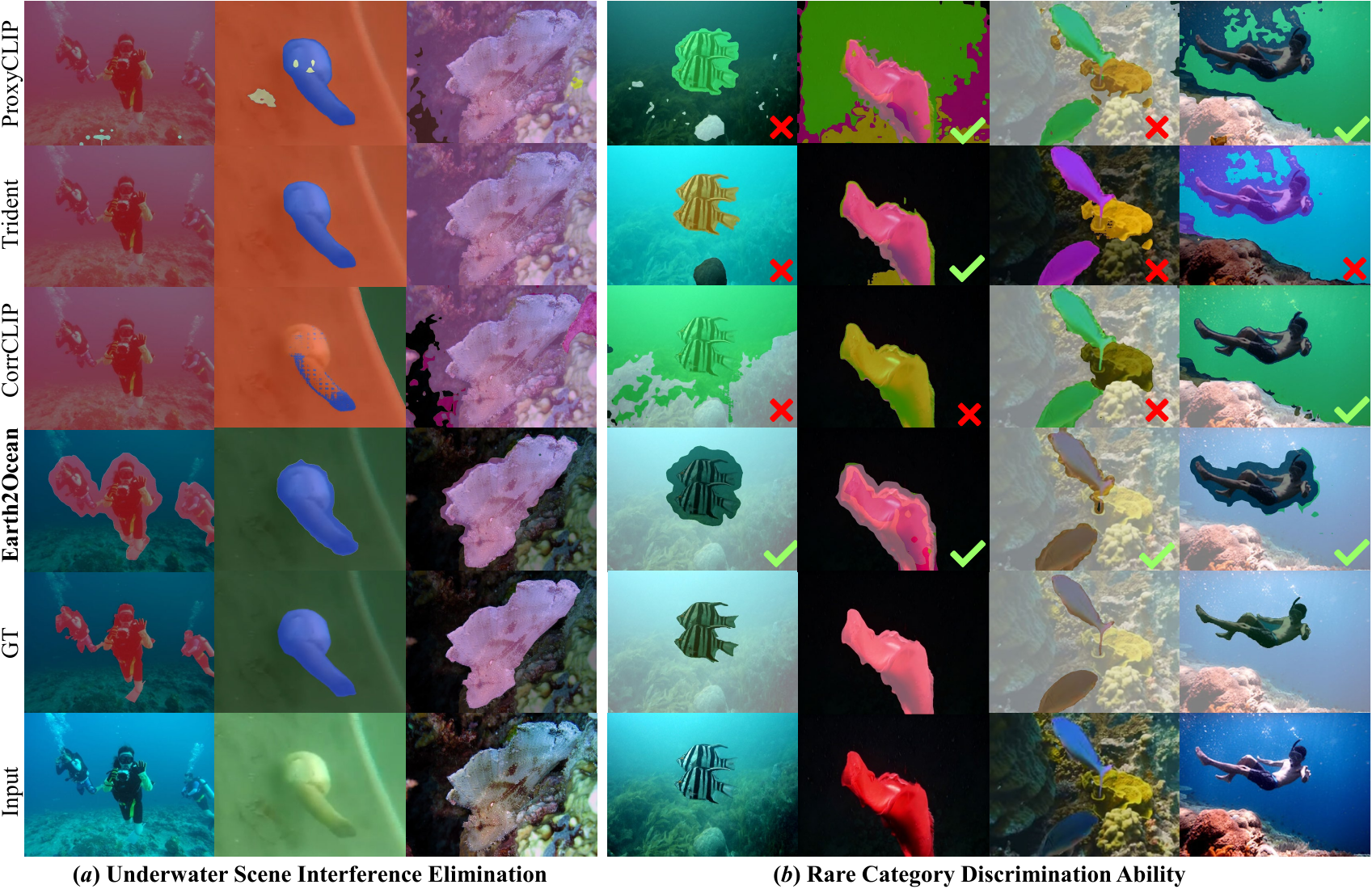}
\vspace{-1.5em}
\caption{(a) Visualization of underwater interference elimination. (b) Visualization of category discrimination ability. 
\protect\includegraphics[height=1em]{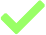} indicates correct classification, while 
\protect\includegraphics[height=1em]{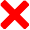} represents misclassification.}
\label{fig:pic_vis_segmap}
\vspace{-1.5em}
\end{figure}

\paragraph{Underwater Scene Interference Elimination.} 
As illustrated in \cref{fig:pic_vis_segmap}(a), our model effectively suppresses underwater scene interferences and restores clear object boundaries by adaptively enhancing geometric visual cues. In contrast, competing methods exhibit blurred contours or residual background noise, leading to unstable predictions in complex underwater conditions.

\paragraph{Category Discrimination Ability.} 
To evaluate category-level discriminability of CSA module, we visualize prediction maps of representative categories, including both common and rare classes. As shown in \cref{fig:pic_vis_segmap}(b), our model exhibits strong category separation, accurately localizing semantic regions with distinct visual cues. Competing approaches, however, tend to confuse rare categories. These results indicate that our model achieves robust category discrimination.

%% file: table/0_MainTable_2.tex
\begin{table*}[htbp]
\centering
\small
\renewcommand{\arraystretch}{1.2}
\setlength{\tabcolsep}{4pt}
\caption{\textbf{Comparison of mIoU and mAcc across grouped categories on the AquaOV255 dataset.} 
The 255 categories in the dataset are grouped by taxonomy and commonness (more split and evaluation details are in the Appendix). 
Abbreviations: \textit{AO = Artificial Objects, IN = Invertebrates, HM = Humans \& Mammals, 
PC = Plants \& Corals, OV = Other Vertebrates.}}
\label{tab:0_MainTable_2_categories}
\resizebox{0.85\linewidth}{!}{%
\begin{tabular}{c|c|c|
cc|cc|cc|cc|cc|cc|
cc|cc|cc|
cc}
\toprule
\multirow{2}{*}{\textbf{Backbone}} & \multirow{2}{*}{\textbf{Method}} & \multirow{2}{*}{\textbf{Publication}} 
& \multicolumn{2}{c}{\cellcolor{blue!5}AO} 
& \multicolumn{2}{c}{\cellcolor{blue!5}Fish} 
& \multicolumn{2}{c}{\cellcolor{blue!5}IN} 
& \multicolumn{2}{c}{\cellcolor{blue!5}HM} 
& \multicolumn{2}{c}{\cellcolor{blue!5}PC} 
& \multicolumn{2}{c|}{\cellcolor{blue!5}OV} 
& \multicolumn{2}{c}{\cellcolor{green!5}Common} 
& \multicolumn{2}{c}{\cellcolor{green!5}General} 
& \multicolumn{2}{c|}{\cellcolor{green!5}Special} 
& \multicolumn{2}{c}{\cellcolor{orange!5}Average} \\
\cmidrule(lr){4-23}
 & & & mIoU & mAcc & mIoU & mAcc & mIoU & mAcc & mIoU & mAcc & mIoU & mAcc & mIoU & mAcc
 & mIoU & mAcc & mIoU & mAcc & mIoU & mAcc & mIoU & mAcc \\
\midrule

\multirow{6}{*}{ViT-B/16} 
 & ClearCLIP & \textit{\textcolor{gray!60}{ECCV2024}} & 21.47 & 27.35 & 8.24 & 14.76 & 6.14 & 11.63 & 35.44 & 51.02 & 9.95 & 42.78 & 45.82 & 46.64 & 21.84 & 32.28 & 11.58 & 18.31 & 7.73 & 13.56 & 11.29 & 18.16 \\
 & SCLIP     & \textit{\textcolor{gray!60}{ECCV2024}} & 16.59 & 26.21 & 4.61 & 11.08 & 5.52 & 10.74 & 26.86 & 48.68 & 13.25 & 29.11 & 36.26 & 52.63 & 16.81 & 32.17 & 9.12 & 17.61 & 4.47 & 9.00 & 8.03 & 15.61 \\
 & ProxyCLIP & \textit{\textcolor{gray!60}{ECCV2024}} & 25.83 & 34.56 & 7.75 & 16.40 & 7.48 & 15.17 & 40.04 & 60.74 & 16.57 & 40.17 & 56.09 & 64.41 & 23.71 & 39.44 & 14.52 & 23.87 & 7.53 & 14.46 & 12.44 & 21.60 \\
 & Trident   & \textit{\textcolor{gray!60}{ICCV2025}} & 28.11 & 36.35 & 9.25 & 18.33 & 7.96 & 16.14 & 48.39 & 69.41 & 15.99 & 40.40 & 60.47 & 65.77 & 26.98 & 42.07 & 16.20 & 25.97 & 8.70 & 16.13 & 14.13 & 23.57 \\
 & CorrCLIP  & \textit{\textcolor{gray!60}{ICCV2025}} & 31.42 & 39.44 & 10.88 & 21.71 & 10.99 & 19.59 & 41.56 & 64.54 & 22.67 & 41.12 & 64.32 & 71.99 & 28.31 & 46.11 & 18.25 & 28.53 & 10.71 & 19.04 & \textcolor{tabblue}{\textbf{16.02}} & \textcolor{tabblue}{\textbf{26.54}} \\
 \rowcolor{gray!10} & Earth2Ocean & \textit{\textcolor{gray!60}{--}} 
 & 36.59 & 46.28 & 10.89 & 19.84 & 14.06 & 24.77 & 45.59 & 65.54 & 31.34 & 59.81 & 68.95 & 82.91 & 30.18 & 46.02 & 20.86 & 32.10 & 11.89 & 19.85 & \textcolor{tabred}{\textbf{17.81}} & \textcolor{tabred}{\textbf{28.06}} \\
\midrule

\multirow{6}{*}{ViT-L/14} 
 & ClearCLIP & \textit{\textcolor{gray!60}{ECCV2024}} & 33.06 & 38.96 & 14.12 & 24.04 & 11.97 & 22.49 & 43.15 & 62.87 & 18.88 & 48.94 & 74.22 & 76.09 & 28.98 & 43.31 & 21.89 & 33.69 & 13.25 & 21.13 & 18.76 & 29.05 \\
 & SCLIP     & \textit{\textcolor{gray!60}{ECCV2024}} & 26.33 & 37.20 & 9.48 & 17.05 & 10.35 & 19.19 & 39.38 & 59.55 & 16.56 & 39.95 & 68.07 & 76.16 & 26.51 & 41.06 & 16.41 & 27.32 & 9.12 & 15.37 & 14.56 & 23.75 \\
 & ProxyCLIP & \textit{\textcolor{gray!60}{ECCV2024}} & 43.30 & 50.24 & 15.36 & 25.66 & 16.02 & 26.79 & 53.91 & 72.49 & 22.92 & 57.18 & 82.09 & 84.85 & 35.13 & 49.41 & 25.62 & 38.85 & 15.81 & 23.78 & 22.29 & 33.03 \\
 & Trident   & \textit{\textcolor{gray!60}{ICCV2025}} & 45.67 & 51.89 & 16.41 & 27.08 & 16.71 & 27.51 & 55.79 & 74.05 & 23.51 & 57.65 & 81.34 & 83.54 & 35.77 & 49.94 & 26.88 & 40.51 & 17.04 & 25.04 & 23.44 & 34.29 \\
 & CorrCLIP  & \textit{\textcolor{gray!60}{ICCV2025}} & 40.27 & 48.84 & 15.88 & 26.49 & 17.60 & 29.20 & 58.66 & 76.28 & 26.42 & 63.94 & 82.29 & 84.94 & 38.99 & 50.98 & 24.18 & 37.96 & 16.25 & 25.75 & \textcolor{tabblue}{\textbf{22.85}} & \textcolor{tabblue}{\textbf{34.12}} \\
  \rowcolor{gray!10} & Earth2Ocean & \textit{\textcolor{gray!60}{--}} 
 & 59.42 & 67.85 & 24.30 & 37.81 & 33.79 & 47.61 & 65.53 & 82.31 & 62.08 & 88.42 & 88.86 & 97.06 & 50.44 & 64.45 & 38.85 & 56.94 & 26.57 & 36.96 & \textcolor{tabred}{\textbf{34.53}} & \textcolor{tabred}{\textbf{47.74}} \\
\midrule

\multirow{6}{*}{ViT-H/14} 
 & ClearCLIP & \textit{\textcolor{gray!60}{ECCV2024}} & 22.63 & 26.51 & 24.39 & 37.16 & 21.54 & 32.72 & 64.31 & 76.73 & 28.31 & 59.56 & 75.79 & 76.66 & 38.43 & 50.46 & 28.48 & 41.93 & 21.24 & 31.33 & 26.42 & 37.93 \\
 & SCLIP     & \textit{\textcolor{gray!60}{ECCV2024}} & 23.65 & 32.22 & 12.84 & 23.16 & 11.91 & 21.36 & 43.92 & 63.82 & 30.37 & 44.73 & 57.24 & 71.45 & 29.27 & 45.05 & 18.87 & 31.10 & 10.94 & 18.71 & 16.58 & 27.14 \\
 & ProxyCLIP & \textit{\textcolor{gray!60}{ECCV2024}} & 34.18 & 40.49 & 22.17 & 35.61 & 19.07 & 31.36 & 59.32 & 74.30 & 51.66 & 72.94 & 82.00 & 88.57 & 40.46 & 56.17 & 29.28 & 43.59 & 19.90 & 30.14 & 26.41 & 38.88 \\
 & Trident   & \textit{\textcolor{gray!60}{ICCV2025}} & 37.29 & 42.78 & 26.34 & 40.55 & 21.97 & 34.83 & 64.79 & 78.51 & 50.22 & 74.04 & 87.38 & 90.76 & 43.80 & 59.40 & 32.87 & 48.32 & 23.89 & 34.31 & 30.17 & 43.04 \\
 & CorrCLIP  & \textit{\textcolor{gray!60}{ICCV2025}} & 39.40 & 44.12 & 24.17 & 37.26 & 23.50 & 35.44 & 62.62 & 78.22 & 68.30 & 85.27 & 84.56 & 91.23 & 43.06 & 57.02 & 32.45 & 46.50 & 23.35 & 33.37 & \textcolor{tabblue}{\textbf{29.57}} & \textcolor{tabblue}{\textbf{41.56}} \\
  \rowcolor{gray!10} & Earth2Ocean & \textit{\textcolor{gray!60}{--}} 
 & 55.04 & 66.03 & 33.81 & 48.17 & 34.33 & 48.46 & 65.47 & 79.19 & 85.46 & 93.98 & 87.09 & 95.55 & 51.95 & 65.96 & 44.00 & 60.14 & 33.52 & 45.78 & \textcolor{tabred}{\textbf{39.76}} & \textcolor{tabred}{\textbf{53.37}} \\
\bottomrule
\end{tabular}%
}
\vspace{-1.5em}
\end{table*}

%% file: table/1_ablation_component.tex
\begin{table}[t]
\centering
\caption{Average performance (aAcc, mIoU, mAcc) of different methods across six underwater segmentation datasets.}
\label{tab:performance_comparison_conponent}
\resizebox{0.9\linewidth}{!}{%
\begin{tabular}{lccc}
\toprule
\textbf{Method} & \textbf{aAcc} & \textbf{mIoU} & \textbf{mAcc} \\
\midrule
Baseline & 34.30 & 15.31 & 25.86 \\
w/ GMG & 46.92 & 27.98 & 42.16 \\
w/ GMG+UWprompt & 49.93 & 30.81 & 46.62 \\
\rowcolor{gray!15}
w/ GMG+UWprompt+CSA & \textcolor{tabred}{\textbf{54.34}} & \textcolor{tabred}{\textbf{34.32}} & \textcolor{tabred}{\textbf{49.84}} \\
\bottomrule
\end{tabular}}
\end{table}

%% file: table/2_ablation_hyper_parameter.tex
\begin{table}[t]
\centering
\small
\caption{\textbf{Ablation study on different hyperparameters.} 
The table shows the effect of various hyperparameters on aAcc, mIoU, and mAcc. 
The best performance for each hyperparameter is highlighted in red bold.}
\label{tab:hyperparam_ablation_transpose}
\resizebox{0.92\linewidth}{!}{
\begin{tabular}{lccccccc}
\toprule
\multirow{4}{*}{(a) $\beta$} & Metric & 0.2 & 0.4 & 0.8 & 1.2 & 1.6 & 3.2 \\
\cmidrule(lr){2-8}
& aAcc & 48.34 & 49.67 & 53.43 & \textcolor{tabred}{\textbf{54.34}} & 52.62 & 6.14 \\
& mIoU & 29.44 & 30.68 & 33.66 & \textcolor{tabred}{\textbf{34.32}} & 32.36 & 1.44 \\
& mAcc & 41.89 & 43.60 & 48.09 & \textcolor{tabred}{\textbf{49.84}} & 49.85 & 9.94 \\
\midrule
\multirow{4}{*}{(b) $\gamma$} & Metric & 1.0 & 2.0 & 3.0 & 4.0 & 5.0 & 6.0 \\
\cmidrule(lr){2-8}
& aAcc & 54.28 & \textcolor{tabred}{\textbf{54.41}} & 54.34 & 54.23 & 54.09 & 53.90 \\
& mIoU & \textcolor{tabred}{\textbf{34.44}} & 34.40 & 34.32 & 34.20 & 34.05 & 33.87 \\
& mAcc & 49.82 & \textcolor{tabred}{\textbf{49.85}} & 49.84 & 49.79 & 49.71 & 49.60 \\
\midrule
\multirow{4}{*}{(c) $w_\text{max}$} & Metric & 0.1 & 0.2 & 0.3 & 0.5 & 0.8 & 1.0 \\
\cmidrule(lr){2-8}
& aAcc & 51.57 & 52.68 & 53.50 & 54.34 & \textcolor{tabred}{\textbf{54.65}} & 54.60 \\
& mIoU & 32.08 & 33.02 & 33.63 & 34.32 & \textcolor{tabred}{\textbf{34.62}} & 34.57 \\
& mAcc & 47.86 & 48.73 & 49.28 & 49.84 & \textcolor{tabred}{\textbf{50.01}} & 49.96 \\
\midrule
\multirow{4}{*}{(d) $\tau$} & Metric & 0.1 & 0.2 & 0.3 & 0.5 & 0.8 & 1.0 \\
\cmidrule(lr){2-8}
& aAcc & \textcolor{tabred}{\textbf{54.35}} & 54.35 & 54.35 & 54.33 & 53.55 & 49.93 \\
& mIoU & \textcolor{tabred}{\textbf{34.33}} & 34.33 & 34.33 & 34.32 & 33.47 & 30.81 \\
& mAcc & \textcolor{tabred}{\textbf{49.86}} & 49.86 & 49.86 & 49.84 & 49.10 & 46.62 \\
\bottomrule
\end{tabular}
}
\vspace{-1.5em}
\end{table}

%% file: table/4_analysis_mllms.tex
\begin{table}[htbp]
\centering
\small
\renewcommand{\arraystretch}{1.0}
\setlength{\tabcolsep}{6pt}
\caption{Analysis of the impact of different MLLMs on Earth2Ocean (ViT-B/16).}
\label{tab:Earth2Ocean_ViTB_avg}
\resizebox{=1.0\linewidth}{!}{
\begin{tabular}{lccc}
\toprule
\textbf{Method} & \textbf{aAcc} & \textbf{mIoU} & \textbf{mAcc} \\
\midrule
\rowcolor{gray!10}
Earth2Ocean(GPT-4o) & \textcolor{tabred}{\textbf{54.34}} & \textcolor{tabred}{\textbf{34.32}} & \textcolor{tabred}{\textbf{49.84}} \\
Earth2Ocean(Qwen2.5VL-3B) & 51.83 & 32.64 & 48.30 \\
Earth2Ocean(Qwen2.5VL-7B) & 52.81 & 33.35 & 47.48 \\
\bottomrule
\end{tabular}
}
\vspace{-1.0em}
\end{table} 

%% file: table/3_analysis_memory_time.tex
\begin{table}[t]
\centering
\small
\caption{\textbf{Performance and complexity comparison.} Memory (MB), FLOPs (G), parameters (M), and segmentation accuracy metrics (aAcc, mIoU, mAcc). \textbf{FPS} is calculated as $1/\text{Time (s)}$. We use \texttt{thop} toolbox for FLOPs calculation.}
\label{tab:flops_comparison_fps}
\resizebox{\linewidth}{!}{
\begin{tabular}{l|cccc|ccc}
\toprule
\textbf{Method} & \textbf{\makecell[c]{Memory\\(MB)}} & \textbf{\makecell[c]{FLOPs\\(G)}} & \textbf{\makecell[c]{Params\\(M)}} & \textbf{FPS} & \textbf{aAcc} & \textbf{mIoU} & \textbf{mAcc} \\
\midrule
ClearCLIP& 573.67 & 269.47 & 57.26 & 90.91 & 38.08 & 18.23 & 30.26 \\
SCLIP& 783.83 & 361.93 & 52.54& 43.48 & 40.94 & 21.57 & 37.55 \\
ProxyCLIP& 1032.17 & 4164.17 & 137.74 & 15.38 & 45.30 & 26.29 & 40.19 \\
Trident& 2623.50 & 2451.09 & 224.17 & 8.77 & 45.70 & 27.14 & 40.52 \\
CorrCLIP& 2610.40 & 4217.29 & 142.46 & 5.68 & 48.32 & 29.56 & 44.21 \\
\midrule
Earth2Ocean& 1032.50 & 2167.48 & 148.64 & 17.54 & \textcolor{tabred}{\textbf{54.34}} & \textcolor{tabred}{\textbf{34.32}} & \textcolor{tabred}{\textbf{49.84}} \\
\bottomrule
\end{tabular}}
\end{table}

%% file: sec/5_conclusion.tex
\section{Conclusion and Limitations}
\label{sec:conclusion}
We present \textbf{AquaOV255} and \textbf{UOVSBench}, fine-grained underwater dataset \& benchmark for OVS, along with \textbf{Earth2Ocean}, a training-free framework that transfers terrestrial VLMs to underwater scenarios. By integrating geometry-guided visual masks (\textbf{GMG}) and category-visual semantic alignment (\textbf{CSA}) via MLLM reasoning, Earth2Ocean achieves accurate mask generation and category-pixel predictions without underwater training. Experiments on six datasets demonstrate significant performance improvements and efficient inference.
Nonetheless, Earth2Ocean still has certain limitations. As it relies on terrestrial VLMs, its ability to segment extremely rare or visually ambiguous underwater species remains constrained. \textbf{Future work} could pre-train underwater VLMs on large-scale underwater datasets to further improve training-free segmentation capability.